\documentclass[conference]{IEEEtran}
\IEEEoverridecommandlockouts
\usepackage{cite}
\usepackage{microtype}
\usepackage{booktabs} 
\pdfoutput=1
\usepackage{amsmath,amssymb,amsfonts}
\usepackage{algorithmic}
\usepackage{graphicx}
\usepackage{textcomp}
\usepackage{xcolor}
\usepackage{hyperref}

\newcommand{\bR}{\mathbf{R}}

\newcommand{\bX}{\mathbf{X}}

\newcommand{\bx}{\mathbf{x}}

\newcommand{\bz}{\mathbf{z}}

\newcommand{\cA}{\mathcal{A}}
\newcommand{\cB}{\mathcal{B}}
\newtheorem{theorem}{Theorem}

\newtheorem{definition}[theorem]{Definition} 

\usepackage{multirow}
\usepackage{enumerate}
\usepackage{color}
\usepackage{subfigure}
\def\BibTeX{{\rm B\kern-.05em{\sc i\kern-.025em b}\kern-.08em
    T\kern-.1667em\lower.7ex\hbox{E}\kern-.125emX}}
\begin{document}

\title{One-shot Learning with Absolute Generalization}

\author{\IEEEauthorblockN{1\textsuperscript{st} Hao Su}
\IEEEauthorblockA{\textit{School of Information and Electronics} \\
\textit{Beijing Institute of Technology}\\
Beijing, China \\
suhao\_hao@126.com}
}

\maketitle

\begin{abstract}
	One-shot learning is proposed to make a pretrained classifier workable 
	on a new dataset based on one labeled samples from each pattern. 
	However, few of researchers consider whether the dataset itself supports 
	one-shot learning. In this paper, we propose a set of definitions to 
	explain what kind of datasets can support one-shot learning and propose 
	the concept
	``absolute generalization". Based on these definitions, we proposed a method 
	to build an absolutely generalizable classifier. The proposed method concatenates 
	two samples as a new single sample, and converts a classification problem 
	to an identity 
	identification problem or a similarity metric problem. Experiments demonstrate 
	that the proposed method is superior to baseline on one-shot learning datasets 
	and artificial datasets. The code can be get 
	\href{https://github.com/qqsuhao/One-shot-Learning-with-Absolute-Genelization}{https://github.com/qqsuhao/One-shot-Learning-with-Absolute-Genelization}
\end{abstract}

\begin{IEEEkeywords}
One-shot learning, Absolute Genelization
\end{IEEEkeywords}

\section{Introduction}
Deep learning models have achieved great success in image classification 
\cite{krizhevsky2017imagenet} \cite{he2016deep}. However, 
in practical applications, the popularization of deep learning models still suffer 
from a lot of problems. The prominent problem is that deep learning models need 
a large number of labeled samples of training \cite{Fan_2020_CVPR} \cite{Kang_2019_ICCV}, 
and once the scenes are changed, 
previously trained models may not work normally although the scenes change does not 
alter semantics about human cognition. 
For example, if we train a 
model to recognize samples with a certain texture of background, we may need to 
retrain the model once the texture of the background changes into another style. 
That is, the generalization of the model is limited, which makes it difficult 
for the model trained on dataset $\cA$ to work on dataset $\cB$. 
Fortunately, studies related to one-shot learning or few-shot learning (FSL) 
in recent years may provide a solution to the problem \cite{o2019one}.

The one-shot learning was first proposed by \cite{fei2006one} and \cite{fe2003bayesian}, 
aiming to learn characteristics of novel patterns from a few of labeled samples 
and use them for classification \cite{sung2018learning}. If one labeled sample 
is taken from each of the 20 novel patterns for learning, it is called 
20-way one-shot learning \cite{koch2015siamese}. In general,  FSL methods can be 
roughly divided into three categories \cite{wang2020generalizing} \cite{Kadam}: 
(1) making augmentation and regularization based on prior knowledge to avoid over-fitting of the model 
\cite{santoro2016meta} \cite{ziko2020laplacian}; (2) constraining hypothesis space 
based on prior knowledge, including multitask learning \cite{zhang2018fine} 
\cite{chopra2005learning}, embedding learning \cite{vinyals2016matching} 
\cite{triantafillou2017few} and so on \cite{Allen}; (3) optimizing the search 
strategy for parameters in hypothesis spaces based on prior knowledge \cite{finn2017model} 
\cite{caelles2017one}.

Most datasets for one-shot learning are divided into two subsets: 
training set and probe set, 
and none of these patterns of the probe set have ever appeared in training set. 
The trained one-shot learning models are expected to learn through few of labeled probe 
samples and be able to classify other unlabeled probe samples. Therefore, it is regarded 
that one-shot learning models are generalizable, which can work on datasets 
that have never been seen before based on limited prior knowledge. If we regard the 
training set and probe set as two different datasets $\cA$ and $\cB$, respectively, 
then one-shot learning models could be regarded as a general model on both datasets.
Most researchers dedicate to study how to improve the 
generalizability by adapting models themselves.
However, few of them consider the relationship between the distribution of data 
and the generalizability of models. That is, the generalizability of models 
may depend on whether the datasets support models to generalize. For example, 
if datasets $\cA$ and $\cB$ are totally different, those models trained on $\cA$ 
certainly could not work on $\cB$. But if $\cA$ and $\cB$ are semantically identical, 
but distributed differently, those models trained on $\cA$ must could be generalized 
on $\cB$ from the perspective of human cognition.

In this paper, we define an ``absolute generalization" for classification from 
the perspective of distribution of dataset. That is, as long as the distribution of 
dataset $\cA$ and $\cB$ satisfies certain conditions, models trained on the dataset 
$\cA$ will be able to work on dataset $\cB$.
We analyze the distribution of concatenated samples 
coupled by two samples and propose a method to make 
existing methods absolutely generalizable. 
The proposed method is aimed to measure the 
similarity between a pair of samples without any metric measurement. Finally, 
we compare our method with the baseline produced by siamese networks for 
verification 
\cite{chopra2005learning} \cite{koch2015siamese}. 

The rest of paper is organized as follows. 
In the second section, we review some related work. 
In the third section, we make some definitions about ``absolute generalization" and 
propose our method to build a absolutely generalizable classifiers. Next, we apply 
experiments compared with siamese networks as baselines in Section 4. 
Finally, we point out the remaining problems and future research directions.

\section{Related work}
\label{sec:related-work}
\subsection{Image Classification}
Deep learning has achieved great success in image classification \cite{pham2020meta}. 
Most of these classifiers are essentially data fitting depending on labels.
Consider the forerunner of convolution neutral networks for recognition, 
LeNet \cite{lecun1998gradient}, which succeeded on MNIST dataset for classification. 
LeNet fits training data to corresponding one-hot code, where appropriate optimizers, 
structures, active functions and loss functions need to be considered.
However, for example, if we simply reverse the color of input images during probe phase, 
these classifiers may not work normally. Although the simple operation does not disturb 
human cognition, these classifiers may take a fatal blow. It is obvious that weights  
in the hidden layers highly depend on the distribution of training data. Once the 
distribution of probe data is changed, no matter whether the change alters 
semantics of probe data, the classifier will not make a difference.
In a broad sense, we can regard that the generalizability of classifier is low.

\subsection{One-shot learning}
\begin{figure}[htpb]
	\centering
	\includegraphics[width=3in]{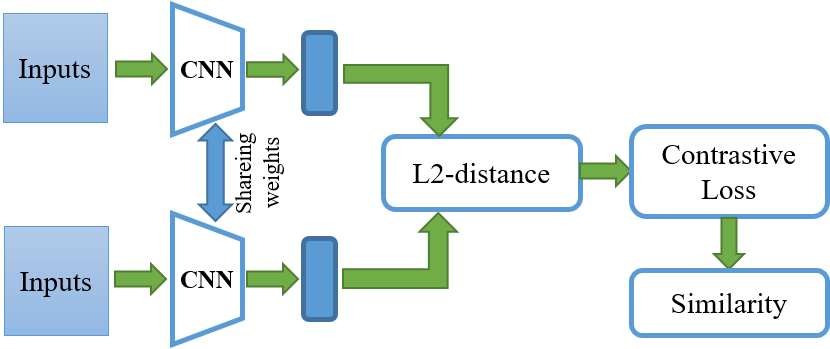}
	\caption{The structure of siamese networks.}
	\label{fig:siamese}
\end{figure}

From the perspective of generalization, these one-shot learning models are aimed to improve 
the generalizability on novel datasets. We can regard a one-shot labeled 
sample as a template. That is, these one-shot learning models recognize unlabeled samples 
through predicting whether they belong to the same pattern with the template 
\cite{o2019one}. Recently, a number of one-shot learning models have been 
developed. In \cite{koch2015siamese}, Koch et al. employed deep siamese networks for 
one-shot learning on Omniglot dataset \cite{sung2018learning}.
The siamese networks were first introduced in the early 1990s by Bromley and LeCun 
to solve signature verification problem \cite{bromley1993signature}, 
which aimed to measure the similarity between a pair of samples.
In \cite{chopra2005learning}, LeCun et al. 
employed two weights-shared subnetworks (usually either CNNs or autoencoders) 
to extract features of a pair samples, respectively, as shown in 
Figure. \ref{fig:siamese}. Then, they 
used a contrastive energy function which contained dual terms to decrease the energy 
of identical pattern pairs and increase the energy of different pattern pairs. 

\section{Methodology}
\label{sec:proposedmethod}

\label{subsec:method}
\begin{figure}[!t]
	\centering
	\includegraphics[width=2.7in]{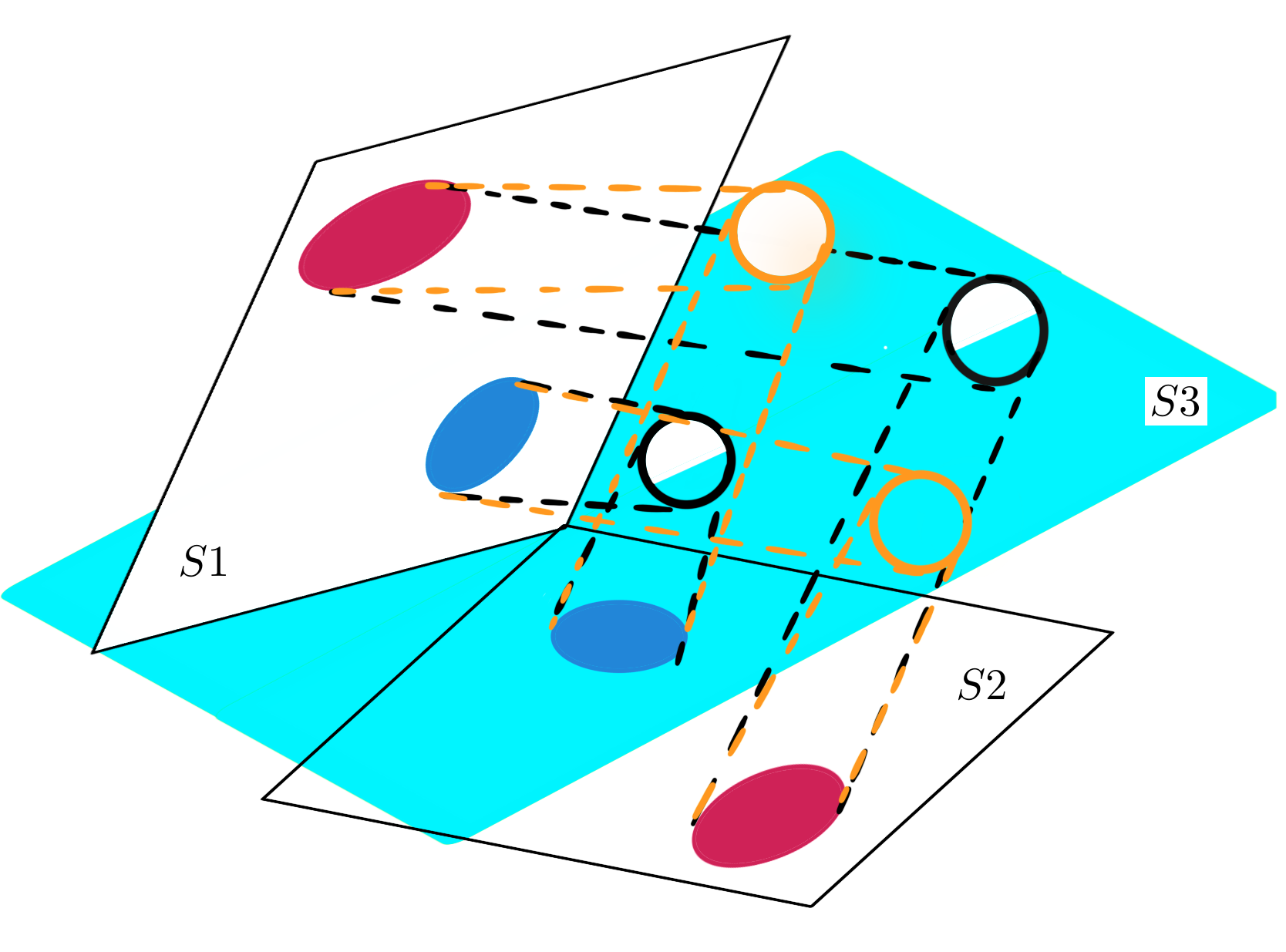}
	\caption{Schematic diagram of our method.}
	\label{fig:example}
\end{figure}

\begin{figure*}[!t]
	\centering
	\subfigure[]{
		\begin{minipage}[t]{0.3\linewidth}
			\centering
			\includegraphics[width=2.25in]{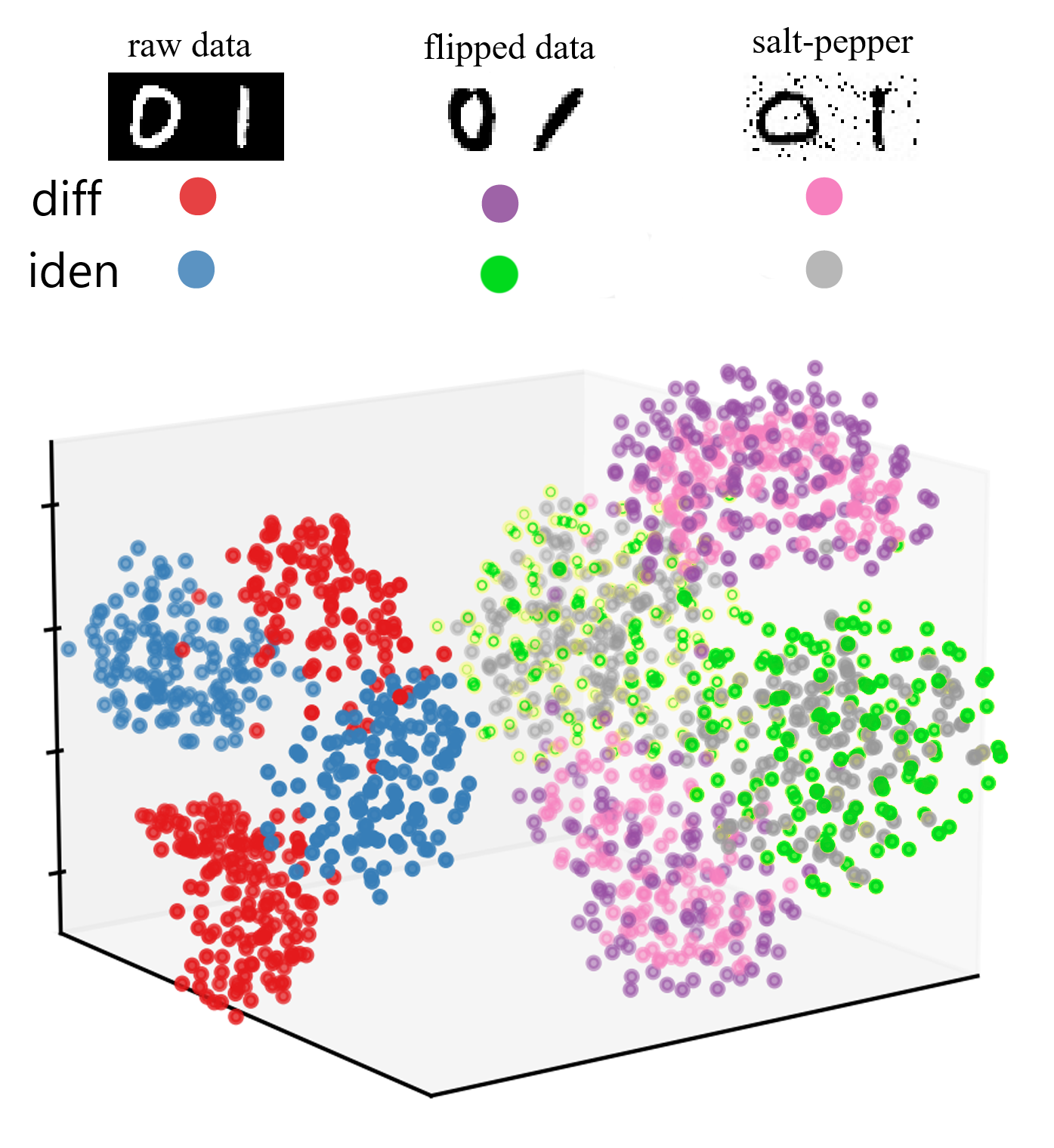}
		\end{minipage}%
	}%
	\subfigure[]{
		\begin{minipage}[t]{0.3\linewidth}
			\centering
			\includegraphics[width=2.2in]{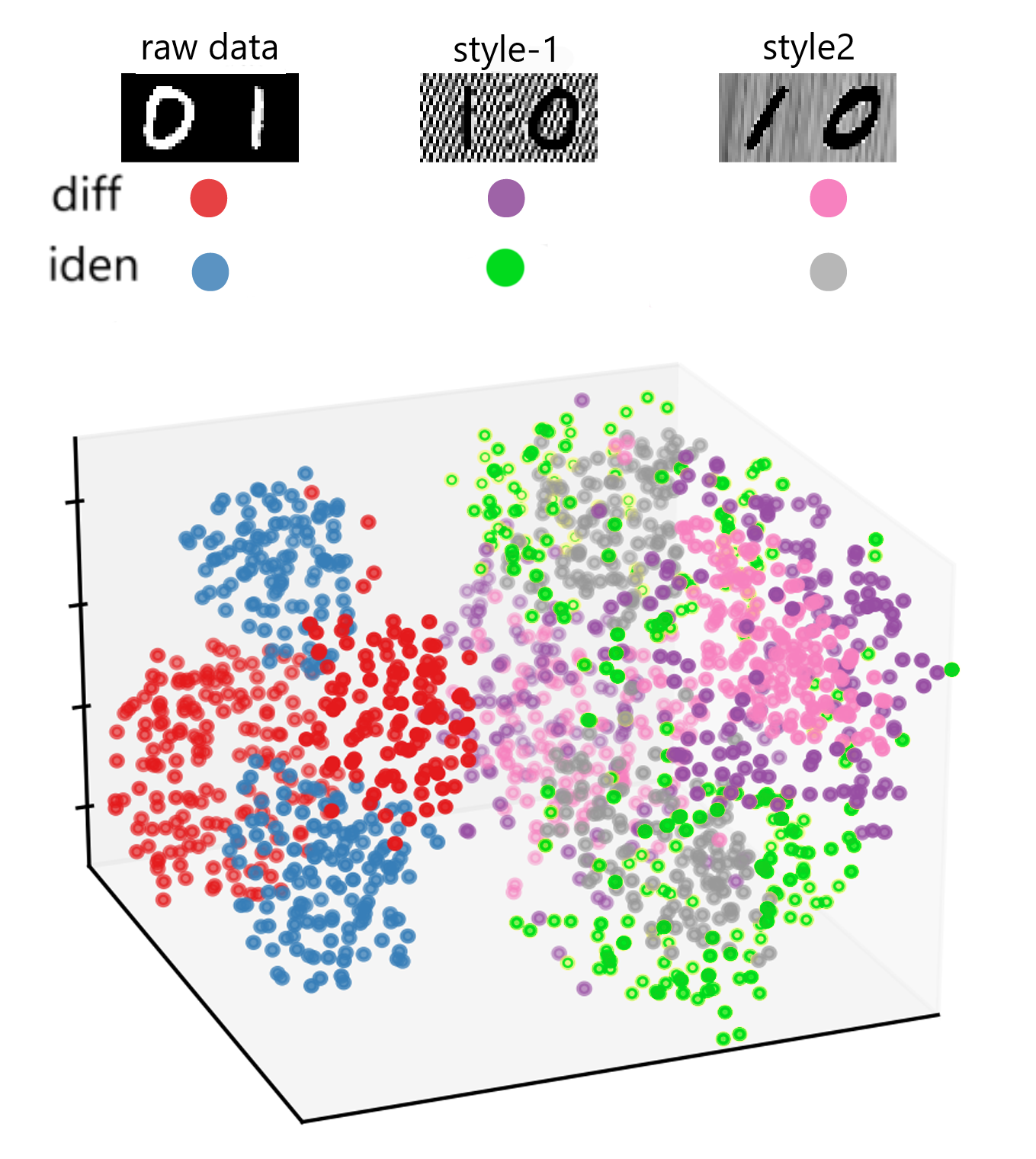}
		\end{minipage}%
	}%
	\subfigure[]{
		\begin{minipage}[t]{0.3\linewidth}
			\centering
			\includegraphics[width=2.04in]{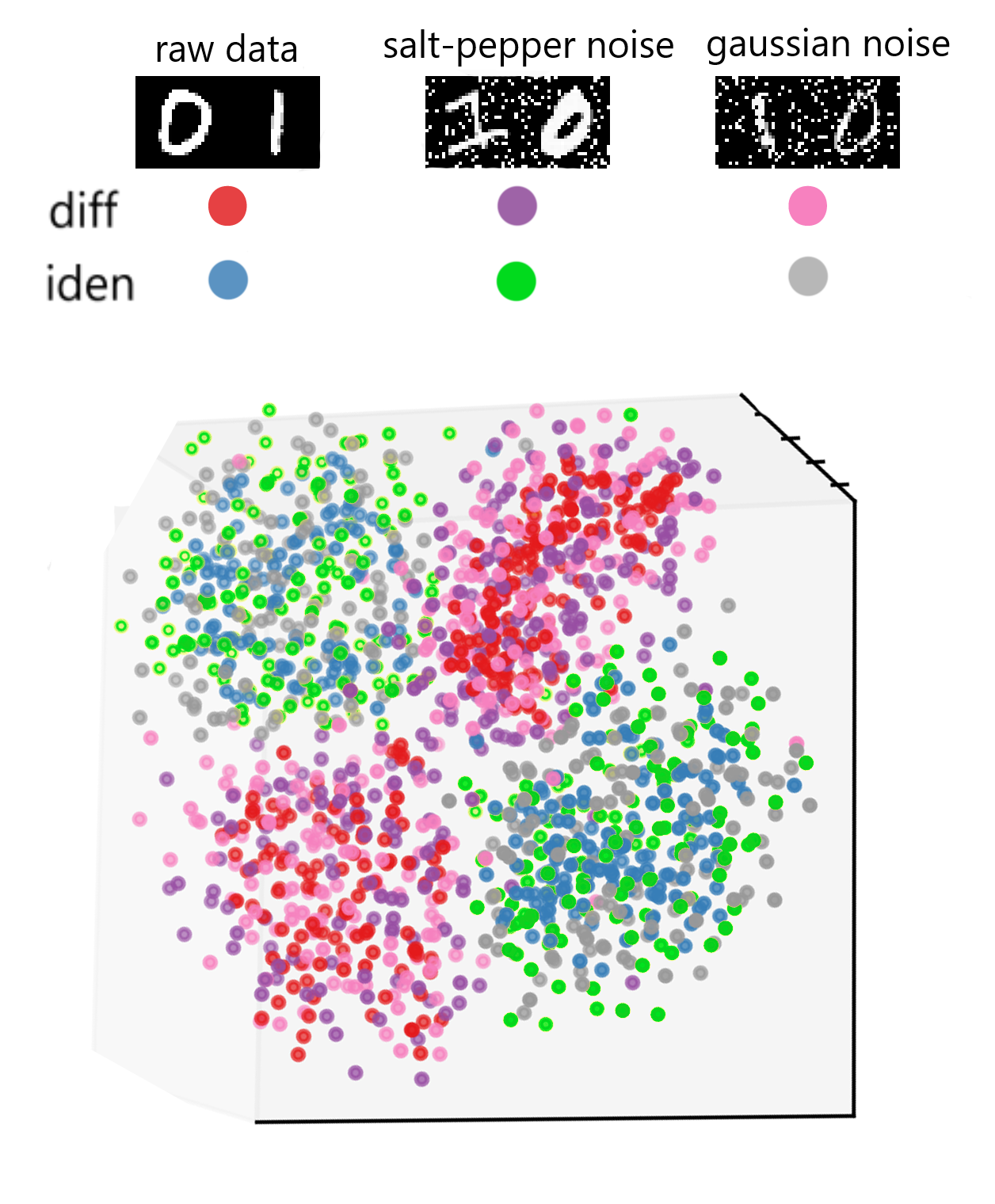}
		\end{minipage}%
	}%
	\caption{Distribution of concatenated samples after reducing dimensions with TSNE.}
	\label{fig:tsne}
\end{figure*}

\subsection{Absolute Generalization}
\label{subsec:AG}
The concept of generalization describes that how well the model works in the probe set
of a dataset. To distinguish the generic concept, we define ``absolute generalization" 
for two datasets $\cA$ and $\cB$:
\begin{definition}
	\label{def:1}
	The absolute generalization refers to that a model which is trained in the
	dataset $\cA$ could be employed on the dataset $\cB$ without extra modification, 
	and we call that the model is absolutely generalizable.
\end{definition}
In above definition, 
the two datasets are different, which is defined as follows:
\begin{definition}
	\label{def:2}
	The dataset $\cA$ is distinguished from the dataset $\cB$ if 
	$f_{\cA}(\bx) \neq f_{\cB}(\bx)$.
\end{definition}
Where, $f_{\cA}(\bx)$ and $f_{\cB}(\bx)$ are data distributions of the 
dataset $\cA$ and $\cB$, respectively. It's noted that the model with absolute 
generalization not always exists for any two datasets. Next, we give the existing 
conditions.

Assume that the dataset $\cA$ contains two patterns which are denoted as $\omega_0$ and 
$\omega_1$, respectively. So does the dataset $\cB$. Generally, we use the maximum 
posterior estimation (MAP) to make decision in the dataset $\cA$:
\begin{equation}
\label{eq:MAP}
\begin{aligned}
\omega_i &= \arg\max\limits_{i}f_{\cA}(\omega_i|\bx)	\\
&= \arg\max\limits_{i}\frac{f_{\cA}(\bx|\omega_i)P_{\cA}(\omega_i)}{f_{\cA}(\bx)}, (i=0, 1) \\
&= \arg\max\limits_{i}f_{\cA}(\bx|\omega_i))P_{\cA}(\omega_i), (i=0, 1)
\end{aligned}
\end{equation}
Where, $P_{\cA}(\omega_i)$ denotes prior probability, which is a constant and could be 
ignored if it is 0.5. Denote $f_{\cA}(\bx|\omega_i)$ as a likelihood distribution 
of the dataset $\cA$. Due to $f_{\cA}(\bx|\omega_i) \neq f_{\cB}(\bx|\omega_i)$ 
and $f_{\cA}(\bx) \neq f_{\cB}(\bx)$, or some models ignore $f_{\cA}(\bx)$,
the classifier trained in the dataset $\cA$ could not be employed in the dataset $\cB$. 
However, we can relax the constraints to make the ``absolute generalization" available.

Assume that the observed sample $x$ in dataset $\cA$ and dataset $\cB$ is generated by the following mapping:
\begin{equation}
\bx = g(\bz_1, \bz_2) 
\end{equation}
Where, denote $g(\cdot)$ as a mapping from latent space to sample space. Denote $\bz_1$ and 
$\bz_2$ as two independent latent variable. As shown in following, we convert the data 
distribution to latent variable distribution:

\begin{align}
\label{eq: condition}
&f_\cA(\bx) \rightarrow f_\cA(\bz_1)f_\cA(\bz_2)	\\
&f_\cA(\bx|\omega_i) \rightarrow f_\cA(\bz_1|\omega_i)f_\cA(\bz_2) \\
&f_\cA(\bz_1) = \sum_{i=0}^{1} f_\cA(\bz_1|\omega_i)P_\cA(\omega_i) \\
&f_\cB(\bx) \rightarrow f_\cB(\bz_1)f_\cB(\bz_2)	\\
&f_\cB(\bx|\omega_i) \rightarrow f_\cB(\bz_1|\omega_i)f_\cB(\bz_2) \\
&f_\cB(\bz_1) = \sum_{i=0}^{1} f_\cB(\bz_1|\omega_i)P_\cB(\omega_i) 
\end{align}
Because the distribution of $\bz_2$ has nothing to do with patterns, we call $f_{\cA}(\bz_2)$ 
as the background distribution of dataset $\cA$.

\begin{definition}
	\label{def:3}
	The classifier with the absolute generalization between two datasets exits 
	if $f_\cA(\bz_1|\omega_i) = f_\cB(\bz_1|\omega_i)$ and 
	$P_\cA(\omega_i) = P_\cB(\omega_i)$, i=0,1. In this case, the classifier is 
	\begin{equation}
	\begin{aligned}
	\omega_i &= \arg\max\limits_{i}\frac{f_{\cA}(\bz_1|\omega_i)P_{\cA}(\omega_i)}{f_{\cA}(\bz_1)} \\
	&=  \arg\max\limits_{i}\frac{f_{\cB}(\bz_1|\omega_i)P_{\cB}(\omega_i)}{f_{\cB}(\bz_1)}
	\end{aligned}
	\end{equation}
\end{definition}
As long as we can isolate the effects of $\bz_2$ when we construct a classifier in the 
dataset $\cA$, the classifier is absolutely generalizable and could be applied in the 
dataset $\cB$.

\subsection{Proposed Method}
Denote sample matrices of dataset $\cA$ and $\cB$ as $\bX_A\in \bR^{m\times N_A}$ 
and $\bX_B\in \bR^{m\times N_B}$, respectively. $m$ denotes dimension of samples. 
$N_A$ and $N_B$ denote number of the samples in dataset, respectively.
Each column of $\bX_A$ is a sample vector and so does $\bX_B$.
Samples of two patterns in dataset $\cA$ are denoted as $\bx^{A_0}_{j}, j=0, \cdots, n_{A_0}$ 
and $\bx^{A_1}_{k}, k=0, \cdots, n_{A_1}$, respectively. Where $n_{A_0}+n_{A_1}=N_{A}$.

Consider converting classification problems of two patterns to distinguishing whether two 
samples belong to an identical pattern or different patterns. 
For dataset $\cA$, 
concatenate samples belonging 
to an identical pattern and denote them as 
$\bx^{iden}_{jk} \in \bX^{A}_{iden} = \{ \begin{bmatrix}
\bx^{A_0}_{j} \\
\bx^{A_0}_{k} \\
\end{bmatrix} | j,k=0,\cdots,n_{A_0} \} \bigcup 
\{ \begin{bmatrix}
\bx^{A_1}_{j} \\
\bx^{A_1}_{k} \\
\end{bmatrix} | j,k=0,\cdots,n_{A_1} \} $ and
concatenate samples belonging to different patterns and denote as 
$\bx^{diff}_{jk} \in \bX^{A}_{diff}=\{
\begin{bmatrix}
\bx^{A_0}_{j} \\
\bx^{A_1}_{k} \\
\end{bmatrix} | j=0,\cdots,n_{A_0};k=0,\cdots,n_{A_1}
\}$. For dataset $\cB$, similar symbols are denoted.
We consider that classifiers in the sample space spanned by $\bX^{A}_{iden}$ and $\bX^{A}_{diff}$ 
are still executable in the sample space spanned by $\bX^{B}_{iden}$ and $\bX^{B}_{diff}$ if 
the two datasets satisfy Definition \ref{def:3}.

We take a simple example to explain the observation. Assume that $m=2$ and the dimension of 
concatenated samples is 4. As shown in Figure. \ref{fig:example}, samples of dataset $\cA$ are 
distributed in 2-dim plane $S1$ with two patterns represented by red and blue, respectively. 
The 2-dim plane $S2$ is a duplicate of $S1$. To be able to show the 4-dim space, regard $S1$ 
and $S2$ as two basis plane. The two orange hollow circles denote concatenate samples 
belonging to different patterns and two black hollow circles denote concatenate samples 
belonging to an identical pattern. Note that two black hollow circle are located in a 
hyperplane passing the origin, denoted as blue plane $S3$. Moreover, two orange hollow circles
are located on both sides of the hyperplane $S3$. In the example, if the two datasets $\cA$ 
and $\cB$ satisfy Definition\ref{def:3}, the slope of decision bound in the two datasets are 
the same, which confirms that the normal vector of hyperplane $S3$ is constant. 
As long as we solve the normal vector in dataset $\cA$, it could be used in dataset $\cB$ for
classification directly.

We use MNIST dataset to further explain our method. We 
concatenate two samples from digits 0 and 1 of MNIST as a new sample and 
generate the sample set $\bX^{A}_{diff}$ and $\bX^{A}_{iden}$. 
We regard the raw sample as 
dataset $\cA$ and $\cA = \bX^{A}_{diff} \bigcup \bX^{A}_{iden}$. 
Then, we modify the distribution of raw data but do not alter its semantics, 
such as flipping image color, adding noise and replacing texture of background. We regard 
these modified dataset as dataset $\cB$. 
Because dataset $\cA$ and $\cB$ have the same semantic, they satisfy Definition 
\ref{def:3}. Due to their different distributions, we regard they are 
two different datasets according to Definition \ref{def:2}.
As shown in Figure. \ref{fig:tsne}, we used t-sne \cite{maaten2008visualizing} 
to reduce the dimension 
of concatenated samples and use different colors to indicate different concatenate 
patterns. For concatenated samples from raw data (i.e. dataset $\cA$) 
denoted as blue and red, the distribution of them is similar to our assumption 
in Figure. \ref{fig:example}. 
The concatenated samples from identical patterns are distributed nearly to the
same hyperplane as other modified datasets in Figure. \ref{fig:tsne}.

%

\subsection{Models for Neural Networks}
\begin{figure}[ht]
	\centering
	\subfigure[Our model with MLP.]{
		\begin{minipage}[t]{\linewidth}
			\centering
			\includegraphics[width=2.2in]{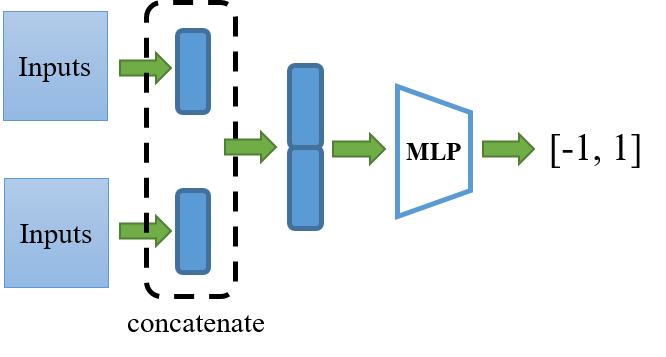}
		\end{minipage}%
	}%
	\\
	\subfigure[Our model with CNN.]{
		\begin{minipage}[t]{\linewidth}
			\centering
			\includegraphics[width=3in]{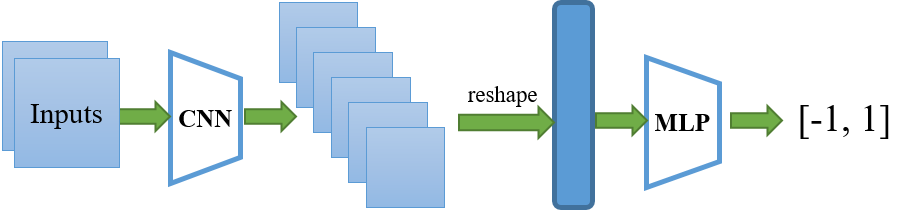}
		\end{minipage}%
	}%
	\caption{Framework of our model.}
	\label{fig:our-model}
\end{figure}

Our method is to couple two samples together as one new sample 
by concatenating them along with a dimension.
In image processing, we directly concatenate two vectors flattened by images 
when employing MLP, as shown in Figure. \ref{fig:our-model}(a). 
When employing CNN, we can concatenate two images along with the channel dimension, 
as shown in Figure. \ref{fig:our-model}(b). 

As explained in Section \ref{subsec:method}, the output of our model represents 
the normalization distance from the concatenated sample to the hyperplane. Thus, 
a positive or negative output indicates that the sample is on both sides of the plane.
We also could regard the output as a probability of 
the concatenated sample belonging to $\bX^{A}_{diff}$, 
if limiting output to $[0, 1]$. If outputs indicate distances, employ 
mean square error (MSE) as loss function; else if they indicate probabilities, 
binary cross entropy (BCE) is employed as loss function.

\subsection{Comparison with Siamese Networks}
Siamese networks \cite{bromley1993signature} and \cite{lecun1998gradient} 
firstly proposed the idea of coupling two samples together, 
where the author employed a distance measurement to 
describe the similarity between two hidden 
features extracted by two identical networks, respectively.
However, in our method, we regard the concatenated sample as a new sample 
and analyze its distribution properties aiming to approve the 
generalizability of classifiers. 

From the perspective of loss function, the siamese networks employ contrastive 
loss function based on p-norm distance. But in our method, it cuold be 
regarded that the type of distance metric is left to the neural 
networks to decide by itself.

The training time and inference time of our model is half that of siamese networks 
because our model only takes one forward propagation.

\section{Experiments}

\subsection{Experiments on MNIST}
\label{sec:MNIST}

\begin{table*}[!t]
	\caption{Experiment on digits 4 and 9 of MNIST.}
	\label{tab1}
	
	\begin{center}
		\begin{small}
			\begin{sc}
				\begin{tabular}{lcccccccccccccr}
					\toprule
					& & \multicolumn{2}{c}{\multirow{2}{*}{raw}} 
					& \multicolumn{2}{c}{\multirow{2}{*}{flipped}} 
					& \multicolumn{2}{c}{salt pepper} 
					& \multicolumn{2}{c}{salt pepper} 
					& \multicolumn{2}{c}{salt pepper} \\
					& & & & & & \multicolumn{2}{c}{noise(0.2)} 
					& \multicolumn{2}{c}{noise(0.5)} & \multicolumn{2}{c}{noise(0.9)}\\
					\midrule
					\multirow{5}{*}{samples} & 
					& \multirow{5}{*}{\includegraphics[scale=0.9]{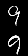}} 
					& \multirow{5}{*}{\includegraphics[scale=0.9]{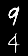}} &
					\multirow{5}{*}{\includegraphics[scale=0.9]{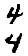}} & 
					\multirow{5}{*}{\includegraphics[scale=0.9]{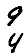}} &
					\multirow{5}{*}{\includegraphics[scale=0.9]{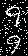}} & 
					\multirow{5}{*}{\includegraphics[scale=0.9]{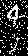}} &
					\multirow{5}{*}{\includegraphics[scale=0.9]{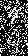}} &
					\multirow{5}{*}{\includegraphics[scale=0.9]{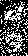}} &
					\multirow{5}{*}{\includegraphics[scale=0.9]{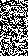}} &
					\multirow{5}{*}{\includegraphics[scale=0.9]{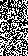}} \\ \\ \\ \\ \\
					\\
					SSIM & & 0.317 & 0.279 & 0.316 & 0.279 & 0.060 & 0.049 & 0.016 & 0.013 & 0.0001 & 0.0001 \\ 
					\\
					\multirow{2}{*}{AUC} & (ours) 
					& \multicolumn{2}{c}{\textbf{0.989}(0.001)} 
					& \multicolumn{2}{c}{\textbf{0.986}(0.002)} 
					& \multicolumn{2}{c}{\textbf{0.951}(0.003)} 
					& \multicolumn{2}{c}{\textbf{0.711}(0.008)} 
					& \multicolumn{2}{c}{\textbf{0.507}(0.004)} \\
					& (SNN) 
					& \multicolumn{2}{c}{0.997(0.001)} 
					& \multicolumn{2}{c}{0.646(0.007)} 
					& \multicolumn{2}{c}{0.901(0.004)} 
					& \multicolumn{2}{c}{0.543(0.006)} 
					& \multicolumn{2}{c}{0.498(0.009)} \\
					& (LeNet) 
					& \multicolumn{2}{c}{0.994(-------)} 
					& \multicolumn{2}{c}{0.725(-------)} 
					& \multicolumn{2}{c}{0.974(-------)} 
					& \multicolumn{2}{c}{0.820(-------)} 
					& \multicolumn{2}{c}{0.517(-------)} \\
					\\
					\multirow{2}{*}{F1-score} & (ours) 
					& \multicolumn{2}{c}{\textbf{0.971}(0.002)} 
					& \multicolumn{2}{c}{\textbf{0.971}(0.003)} 
					& \multicolumn{2}{c}{\textbf{0.895}(0.004)} 
					& \multicolumn{2}{c}{\textbf{0.708}(0.009)} 
					& \multicolumn{2}{c}{\textbf{0.669}(0.006)} \\
					& (SNN) 
					& \multicolumn{2}{c}{0.983(0.001)} 
					& \multicolumn{2}{c}{0.687(0.007)} 
					& \multicolumn{2}{c}{0.837(0.005)} 
					& \multicolumn{2}{c}{0.666(0.008)} 
					& \multicolumn{2}{c}{0.665(0.008)} \\
					& (LeNet) 
					& \multicolumn{2}{c}{0.993(-------)} 
					& \multicolumn{2}{c}{0.784(-------)} 
					& \multicolumn{2}{c}{0.961(-------)} 
					& \multicolumn{2}{c}{0.787(-------)} 
					& \multicolumn{2}{c}{0.673(-------)} \\
					\bottomrule
				\end{tabular}
			\end{sc}
		\end{small}
	\end{center}
	
	\begin{center}
		\begin{small}
			\begin{sc}
				\begin{tabular}{lcccccccccccccr}
					\toprule
					& & \multicolumn{2}{c}{gaussian}
					& \multicolumn{2}{c}{gaussian}
					& \multicolumn{2}{c}{gaussian} 
					& \multicolumn{2}{c}{\multirow{2}{*}{style 1}} 
					& \multicolumn{2}{c}{\multirow{2}{*}{style 2}} \\
					& & \multicolumn{2}{c}{noise(0.5)} 
					& \multicolumn{2}{c}{noise(0.9)} 
					& \multicolumn{2}{c}{noise(1.5)} \\
					\midrule
					\multirow{5}{*}{samples} & & 
					\multirow{5}{*}{\includegraphics[scale=0.9]{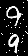}} & 
					\multirow{5}{*}{\includegraphics[scale=0.9]{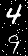}} &
					\multirow{5}{*}{\includegraphics[scale=0.9]{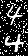}} & 
					\multirow{5}{*}{\includegraphics[scale=0.9]{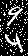}} &
					\multirow{5}{*}{\includegraphics[scale=0.9]{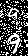}} & 
					\multirow{5}{*}{\includegraphics[scale=0.9]{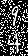}} &
					\multirow{5}{*}{\includegraphics[scale=0.9]{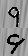}} &
					\multirow{5}{*}{\includegraphics[scale=0.9]{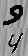}} &
					\multirow{5}{*}{\includegraphics[scale=0.9]{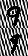}} &
					\multirow{5}{*}{\includegraphics[scale=0.9]{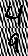}} \\ \\ \\ \\ \\
					\\
					SSIM & & 0.140 & 0.114 & 0.053 & 0.039& 0.021 & 0.016 & 0.167 & 0.142 & 0.060 & 0.046 \\ 
					\\
					\multirow{2}{*}{AUC} & (ours) 
					& \multicolumn{2}{c}{\textbf{0.988}(0.001)} 
					& \multicolumn{2}{c}{\textbf{0.954}(0.003)} 
					& \multicolumn{2}{c}{\textbf{0.897}(0.006)} 
					& \multicolumn{2}{c}{\textbf{0.984}(0.002)} 
					& \multicolumn{2}{c}{\textbf{0.973}(0.002)} \\
					& (SNN) 
					& \multicolumn{2}{c}{0.993(0.001)} 
					& \multicolumn{2}{c}{0.875(0.003)} 
					& \multicolumn{2}{c}{0.701(0.006)} 
					& \multicolumn{2}{c}{0.593(0.007)} 
					& \multicolumn{2}{c}{0.544(0.008)} \\
					& (LeNet) 
					& \multicolumn{2}{c}{0.994(-------)} 
					& \multicolumn{2}{c}{0.986(-------)} 
					& \multicolumn{2}{c}{0.971(-------)} 
					& \multicolumn{2}{c}{0.663(-------)} 
					& \multicolumn{2}{c}{0.694(-------)} \\
					\\
					\multirow{2}{*}{F1-score} & (ours) 
					& \multicolumn{2}{c}{\textbf{0.965}(0.002)} 
					& \multicolumn{2}{c}{\textbf{0.896}(0.004)} 
					& \multicolumn{2}{c}{\textbf{0.828}(0.005)} 
					& \multicolumn{2}{c}{\textbf{0.953}(0.003)} 
					& \multicolumn{2}{c}{\textbf{0.924}(0.004)} \\
					& (SNN) 
					& \multicolumn{2}{c}{0.973(0.002)} 
					& \multicolumn{2}{c}{0.811(0.007)} 
					& \multicolumn{2}{c}{0.691(0.006)}
					& \multicolumn{2}{c}{0.677(0.005)} 
					& \multicolumn{2}{c}{0.674(0.007)} \\
					& (LeNet) 
					& \multicolumn{2}{c}{0.991(-------)} 
					& \multicolumn{2}{c}{0.970(-------)} 
					& \multicolumn{2}{c}{0.937(-------)} 
					& \multicolumn{2}{c}{0.749(-------)} 
					& \multicolumn{2}{c}{0.759(-------)} \\
					\bottomrule
				\end{tabular}
			\end{sc}
		\end{small}
		\vskip 0.05in
		\footnotesize{Note that the ``salt pepper noise(0.2)" denotes the density 
			of salt pepper noise is 0.2. }
		
		\footnotesize{The ``gaussian noise(0.5)" denotes the variance of gaussian 
			noise is 0.5. So do others.}
	\end{center}
	
\end{table*}

\begin{table*}[ht]
	\caption{Experiments on ORL face datasets.}
	\label{tab2}
	
	\begin{center}
		\begin{small}
			\begin{sc}
				\resizebox{\textwidth}{1.3in}{
					\begin{tabular}{lcccccccccccr}
						\toprule
						& & \multicolumn{2}{c}{\multirow{2}{*}{raw}} 
						& \multicolumn{2}{c}{\multirow{2}{*}{flipped}} 
						& \multicolumn{2}{c}{salt pepper}  
						& \multicolumn{2}{c}{gaussian} \\
						& & & & & & \multicolumn{2}{c}{noise(0.1)} &  \multicolumn{2}{c}{noise(50)}\\
						\midrule
						\multirow{7}{*}{samples} & 
						& \multirow{7}{*}{\includegraphics[scale=0.3]{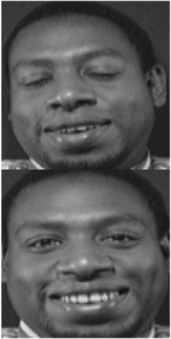}} 
						& \multirow{7}{*}{\includegraphics[scale=0.3]{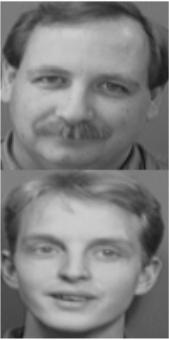}} &
						\multirow{7}{*}{\includegraphics[scale=0.3]{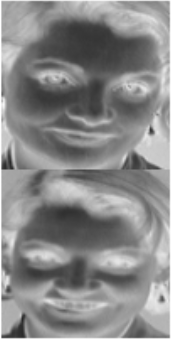}} & 
						\multirow{7}{*}{\includegraphics[scale=0.3]{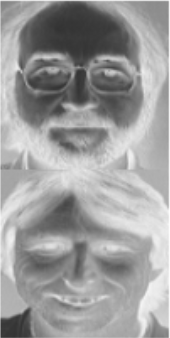}} &
						\multirow{7}{*}{\includegraphics[scale=0.3]{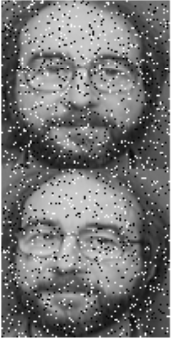}} & 
						\multirow{7}{*}{\includegraphics[scale=0.3]{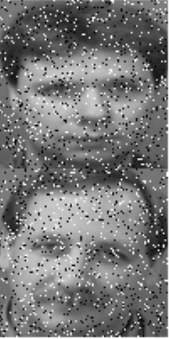}} &
						\multirow{7}{*}{\includegraphics[scale=0.3]{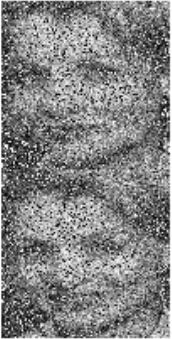}} &
						\multirow{7}{*}{\includegraphics[scale=0.3]{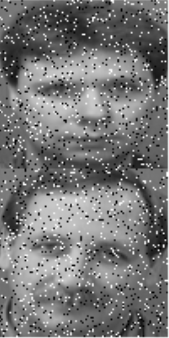}} \\ \\ \\ \\ \\ \\ \\ \\
					\\
						SSIM & & 0.575 & 0.385 & 0.568 & 0.392 & 0.096 & 0.065 & 0.047 & 0.037 \\ 
					\\
						\multirow{2}{*}{AUC} & (ours) & \multicolumn{2}{c}{\textbf{0.943}(0.006)} 
						& \multicolumn{2}{c}{\textbf{0.910}(0.010)} 
						& \multicolumn{2}{c}{\textbf{0.935}(0.007)}
						& \multicolumn{2}{c}{\textbf{0.894}(0.005)} \\
						& (SNN) & \multicolumn{2}{c}{0.943(0.007)} 
						& \multicolumn{2}{c}{0.844(0.019)} 
						& \multicolumn{2}{c}{0.509(0.012)} 
						& \multicolumn{2}{c}{0.606(0.014)} \\
					\\
						\multirow{2}{*}{F1-score} & (ours) 
						& \multicolumn{2}{c}{\textbf{0.900}(0.005)} 
						& \multicolumn{2}{c}{\textbf{0.866}(0.009)} 
						& \multicolumn{2}{c}{\textbf{0.875}(0.005)} 
						& \multicolumn{2}{c}{\textbf{0.821}(0.009)} \\
						& (SNN) & \multicolumn{2}{c}{0.882(0.016)} & \multicolumn{2}{c}{0.794(0.018)} 
						& \multicolumn{2}{c}{0.665(0.018)}
						& \multicolumn{2}{c}{0.677(0.021)} \\
						\bottomrule
				\end{tabular}  }
			\end{sc}
		\end{small}
	\end{center}
\end{table*}

\begin{table*}[ht]
	\caption{Experiments on Omniglot datasets.}
	\label{tab3}
	
	\begin{center}
		\begin{small}
			\begin{sc}
				\resizebox{\textwidth}{1.1in}{
					\begin{tabular}{lccccccccccccccr}
						\toprule
						& \multicolumn{2}{c}{\multirow{2}{*}{raw}} 
						& \multicolumn{2}{c}{\multirow{2}{*}{flipped}} 
						& \multicolumn{2}{c}{salt pepper}  
						& \multicolumn{2}{c}{gaussian}
						& \multicolumn{2}{c}{\multirow{2}{*}{style1}}
						& \multicolumn{2}{c}{\multirow{2}{*}{style2}}
						\\
						& & & & & \multicolumn{2}{c}{noise(0.5)} &  \multicolumn{2}{c}{noise(0.9)}\\
						\midrule
						& 
						\multicolumn{2}{c}{\multirow{5}{*}{\includegraphics[scale=0.3]{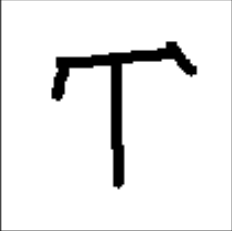}}} & 
						\multicolumn{2}{c}{\multirow{5}{*}{\includegraphics[scale=0.3]{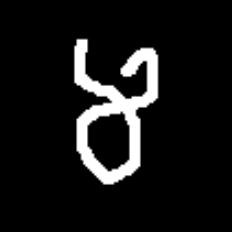}}} & 
						\multicolumn{2}{c}{\multirow{5}{*}{\includegraphics[scale=0.3]{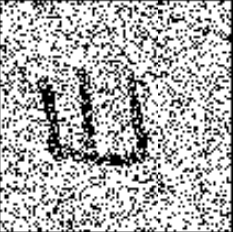}}} & 
						\multicolumn{2}{c}{\multirow{5}{*}{\includegraphics[scale=0.3]{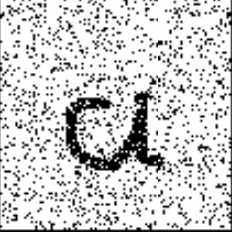}}} &
						\multicolumn{2}{c}{\multirow{5}{*}{\includegraphics[scale=0.3]{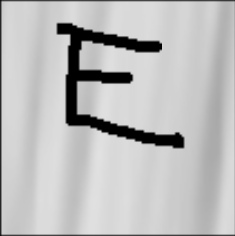}}} &
						\multicolumn{2}{c}{\multirow{5}{*}{\includegraphics[scale=0.3]{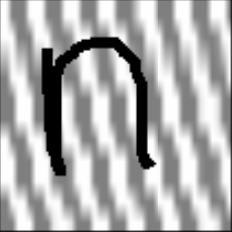}}} &
						\\ \\ \\ \\ \\ \\ 
					\\
						1-shot & 5-way & 20-way & 5-way & 20-way & 5-way & 20-way & 5-way & 20-way & 5-way & 20-way 
						& 5-way & 20-way \\
					\\
						\multirow{2}{*}{(ours)} & \textbf{0.974} & \textbf{0.916} & 0.427 & 
						0.190 & \textbf{0.823} & \textbf{0.581} 
						& \textbf{0.958} & \textbf{0.861} & \textbf{0.970} & \textbf{0.906} & \textbf{0.956} & \textbf{0.879} \\
						& (0.006) & (0.014) & (0.028) & (0.021) & (0.015) & (0.017) 
						& (0.006) & (0.018) & (0.007) & (0.016) & (0.011) & (0.014)\\ \\
						\multirow{2}{*}{(SNN)} & 0.968 & 0.896 & 0.661 & 0.380 & 0.743 & 0.495 
						& 0.942 & 0.813 & 0.964 & 0.887 & 0.959 & 0.867 \\
						& (0.006) & (0.018) & (0.028) & (0.027) & (0.023) & (0.030) 
						& (0.010) & (0.021) & (0.012) & (0.015) & (0.010) & (0.014) \\
						\bottomrule
				\end{tabular}  }
			\end{sc}
		\end{small}
	\end{center} 
\end{table*}

\begin{figure}[ht]
	\centering
	\includegraphics[width=3.2in]{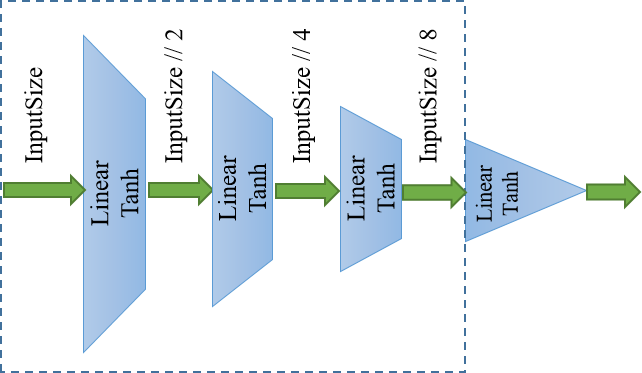}
	\caption{The structure of MLP employed in our model and SNN.}
	\label{fig:MLP}
\end{figure}

The MNIST dataset \cite{lecun1998gradient} contains 70,000 samples of 10 patterns. Each sample is
a single channel image of $28*28$ with black background and white foreground. Due to 
our attention on ``absolute generalization", we only use the digit 4 and 9 in our experiment.
We train our model on the raw dataset and test the model on some modified samples to valid 
the performance of our model. For comparison, we employ siamese neutral networks 
(SNN) as a contrast.

Fairly, the same structure of MLP is employed in our model and SNN. As shown 
in Figure. \ref{fig:MLP}, the whole structure is employed in our model and the first 4 layers are employed in SNN as a subnetwork that shares weights. 
The image samples are reshaped as vectors at the input layer of the network.
The variables' dimension drops by half as they pass through a full connection layer. 
Because that the output of Tanh is in block $[-1, 1]$, we use Tanh as activate layers 
according to \ref{fig:our-model}.

In training phase, we train models with 100 epoches. For each epoch, 
23582 pairs of samples are chosen randomly in the 
MNIST dataset with batch-size 256. We use Adam optimizer with learning rate 0.001. In probe 
phase, besides raw probe dataset of MNIST, some modified samples are produced 
with noise or different styles based on the raw probe dataset of MNIST, which are 
regarded as another dataset different from the MNIST dataset. As shown in Table. \ref{tab1}, 
we have 10 different probe datasets including raw data, flipped black and white, 
adding salt-pepper noise with various noisy pixel densities, 
adding Gaussian noise with various variances, adding various texture styles on background. We use 
area under the curve (AUC) of the receiver operating characteristics (ROC) and F1-scores 
to measure the classification performance, where F1-score is the harmonic average of precision and recall.
We operate each experiment 10 times and get the mean and standard deviation.
As references, we use average Structural SIMilarity (SSIM) 
\cite{wang2004image} to measure the similarity between pairs of samples.

Besides, we use LeNet trained with digits 4 and 9 of raw training dataset to compared with our model. 
Note that the intput of LeNet is one image, which is different from SNN and ours. 
That is, the datasets LeNet used is different from SNN and ours.
Thus, the AUC and F1-socre of LeNet is just a reference.

As shown in Table. \ref{tab1}, when using simple structure of MLP, 
SNN is awkward in probe datasets 
adding salt-pepper noise and  adding various texture styles on background. 
However, our model is better at these probe dataset. 
The LeNet does a good job on 
probe datasets adding noise. However, it has a poor performance on ``flliped", 
``style1" and ``style2". Due to the input of our model 
is a concatenated sample coupling two images, where one of them provides prior knowledge 
when another is regarded as a probe sample, our model has better performance.

\subsection{Experiments for Face Identification}
\label{sec:orl}
\begin{figure}[ht]
	\centering
	\subfigure[SNN.]{
		\begin{minipage}[t]{0.43\linewidth}
			\centering
			\includegraphics[width=1.3in]{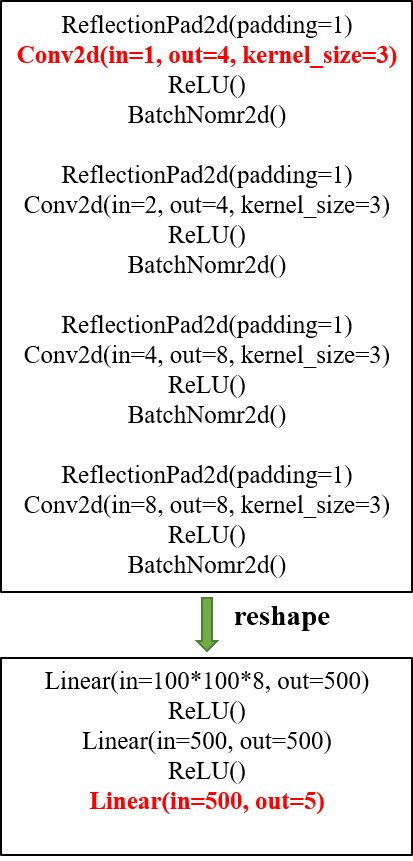}
		\end{minipage}%
	}%
	\subfigure[Our model.]{
		\begin{minipage}[t]{0.43\linewidth}
			\centering
			\includegraphics[width=1.3in]{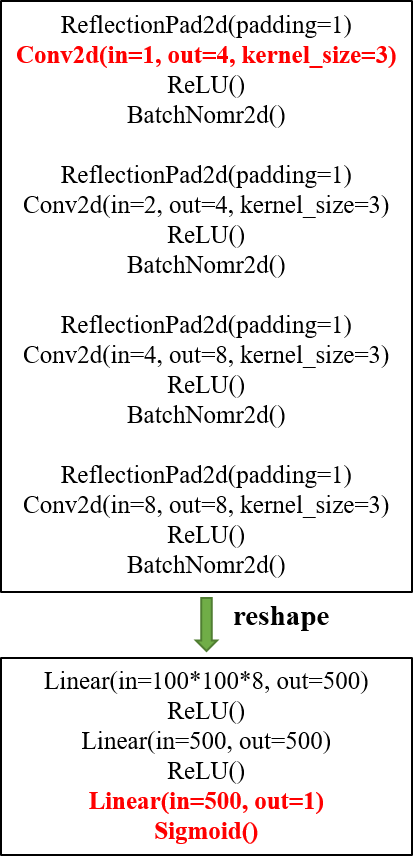}
		\end{minipage}%
	}%
	\caption{Structures of CNN in the experiment on ORL face dataset.}
	\label{fig:CNN}
\end{figure}

We conduct the experiment on ORL \cite{samaria1994parameterisation} faces Dataset. 
The dataset contains 40 patterns of faces with 
10 images in each pattern. All images are stored in grayscale with image size $92\times 112$. 
For each pattern of faces, the images were captured at different times, under different lighting, 
with different facial expressions (eyes open/closed, smiling/not smiling) and facial details 
(with glasses/without glasses). All images were taken against a dark, uniform background, 
with the front face (some slightly sideways).

Fairly, we employ the same convolution neutral networks in our model and SNN, as shown in 
Figure. \ref{fig:CNN}, the difference between two models is the input and output of the network. Note that we use ReLU and Sigmoid as activate layers rather than Tanh. 
Because that the distributions of concatenated samples are complex 
when the patterns in a dataset is more than 2, we only use 0 and 1 to 
denote the label of identical pattern samples and different pattern samples, 
respectively.

In the experiment, we reshape all images to the size of $100\times 100$. In training phase, 
we use 20 patterns of faces to train our model and SNN. We train models with 100 epoches. 
For each epoch, 660 pairs of samples are chosen randomly which are evenly split between 
the identical patterns sample pair and the different patterns sample pair. In probe phase, 
we evaluate models on probe datasets including the rest 20 patterns of faces. Similarly to 
the Section \ref{sec:MNIST}, we produce some images as different datasets 
based on the raw probe dataset.

As shown in Table. \ref{tab2}, SNN has lower recognition rate for probe samples with salt-pepper noise compared with our model.
our model has better performance than SNN in face identification.

\subsection{Experiments on Omniglot for One-Shot Learning}
\label{sec:omniglot}
The Omniglot dataset \cite{lake2011one} is a classical dataset for one-shot learning, 
which is collected 
by Brenden Lake and his collaborators. The dataset contains handwritten character images 
from 50 alphabets ranging from well-established international languages. All images are 
divided into a 40 alphabet background set and a 10 alphabet evaluation set, which are used 
for training and probe phase, respectively.

For fairness, we employ the similar structure in our model and SNN as reference \cite{koch2015siamese}. 
In training phase, no data augmentation is used. In probe phase, we produce 5 datasets based on 
the raw probe dataset, including ``flipped", adding salt-pepper noise with density 0.5, 
adding gaussian noise with variance 0.9 and styles transformation.

As shown in Table. \ref{tab3}, our model perform better than SNN on most probe datasets. 
However, on the ``flipped" dataset, our model perform worse. Theoretically, the 
distribution of ``flipped" dataset is a mirror of raw dataset, which could be classified 
as accurate as raw dataset. However, in Section \ref{sec:orl} and \ref{sec:omniglot}, the 
accuracy on the ``flipped" dataset is far lower than its on the raw dataset, which may be 
because that weights of CNN in our model still tightly depend on the distribution of 
background.

\section{Conclusion}
In practical applications, we always expect that a trained model could 
deal with other similar classification task. One-shot learning gives a 
solution for the multi-task need. However, most researchers focus on 
improving the classification performance of model, few of them consider 
whether the datasets they employed are fit for one-shot learning.

In this paper, we proposed the concept ``absolute generalization" in order 
to explain what kind of datasets were fit for one-shot learning. 
We believed that a classifier with absolute generalizability can be obtained 
when the datasets satisfied certain conditions. 
We proposed a method to build an absolutely generalizable classifier. 
In the method, a new dataset was produced by concatenating two samples of raw 
datasets. In the new dataset, we converted a classification problem to an identity 
identification problem or a similarity metric problem. The distribution of the 
new dataset hid a constant hyperplane which supported an absolutely generalizable classifier.

Because open source datasets cannot satisfy our conditions, we produced 
some artificial datasets based on open source datasets. 
However, these artificial datasets had a great challenge and practical 
significance. Experiments showed that the proposed method was superior 
to the baseline method, which confirmed that our concerns did influence 
the baseline method.

However, we found that the proposed method performed poorly when combined 
with CNN. So in the future, we will continue to study our method based on 
CNN. Besides, we will try to concatenate samples in higher dimensions for 
few-shot learning.

\bibliography{su_ref}
\bibliographystyle{IEEEtran}

\end{document}